\documentclass[manuscript, screen]{acmart}
\def\BibTeX{{\rm B\kern-.05em{\sc i\kern-.025em b}\kern-.08emT\kern-.1667em\lower.7ex\hbox{E}\kern-.125emX}}

\renewcommand\footnotetextcopyrightpermission[1]{}
\usepackage{subcaption}
\usepackage{graphicx}
\newtheorem{mydef}{Definition}

\begin{document}

%
\title{Locally Differentially Private Naive Bayes Classification}

%
\author{Emre Yilmaz}
\email{emre.yilmaz@case.edu}
\affiliation{%
  \institution{Case Western Reserve University}
}

\author{Mohammad Al-Rubaie}
\email{mtilink@gmail.com}
\affiliation{%
  \institution{University of South Florida}
}

\author{J. Morris Chang}
\email{chang5@usf.edu}
\affiliation{%
  \institution{University of South Florida}
%
}

%
\renewcommand{\shortauthors}{Yilmaz, et al.}

%
\begin{abstract}
In machine learning, classification models need to be trained in order to predict class labels. When the training data contains personal information about individuals, collecting training data becomes difficult due to privacy concerns. Local differential privacy is a definition to measure the individual privacy when there is no trusted data curator. Individuals interact with an untrusted data aggregator who obtains statistical information about the population without learning personal data. In order to train a Naive Bayes classifier in an untrusted setting, we propose to use methods satisfying local differential privacy. Individuals send their perturbed inputs that keep the relationship between the feature values and class labels. The data aggregator estimates all probabilities needed by the Naive Bayes classifier. Then, new instances can be classified based on the estimated probabilities. We propose solutions for both discrete and continuous data. In order to eliminate high amount of noise and decrease communication cost in multi-dimensional data, we propose utilizing dimensionality reduction techniques which can be applied by individuals before perturbing their inputs. Our experimental results show that the accuracy of the Naive Bayes classifier is maintained even when the individual privacy is guaranteed under local differential privacy, and that using dimensionality reduction enhances the accuracy.
\end{abstract}

%
%

\ccsdesc[500]{Security and privacy~Privacy-preserving protocols}
\ccsdesc[500]{Computing methodologies~Supervised learning by classification}
\ccsdesc[100]{Computing methodologies~Dimensionality reduction and manifold learning}

%
\keywords{Local Differential Privacy, Naive Bayes, Classification, Dimensionality Reduction}

%

%
\maketitle

\section{Introduction}\label{section:introduction}

Predictive analytics is the process of making prediction about future events by analyzing the current data using statistical techniques. It is used in many different areas such as marketing, insurance, financial services, mobility, and healthcare. For predictive analytics many techniques can be used from statistics, data mining, machine learning, and artificial intelligence. Classification methods in machine learning such as neural networks, support vector machines, regression techniques, and Naive Bayes are widely used for predictive analytics. These methods are supervised learning methods in which labeled training data is used to generate a function which can be used for classifying new instances. In these supervised learning methods, the accuracy of the classifier highly depends on the training data. Using a larger training set improves the accuracy most of the time. Hence, one needs to have a large training data in order to do classification accurately. However, collecting a large dataset brings privacy concerns. In many real life applications, the classification tasks require training sets containing sensitive information about individuals such as financial, medical or location information. For instance, insurance companies need financial information of individuals for risk classification. If there is a company that wants to build a model for risk classification, the data collection may be a critical problem because of privacy concerns. Therefore, we address the problem of doing classification while protecting the privacy of the individuals who provide the training data; thus enabling companies and organizations to achieve their utility targets, while helping individuals to protect their privacy. 

Differential privacy is a commonly used standard for quantifying individual privacy. In the original definition of differential privacy \cite{dwork2008differential}, there is a trusted data curator which collects data from individuals and applies techniques to obtain differentially private statistics about the population. Then, the data curator publishes privacy-preserving statistics about the population. Satisfying differential privacy in the context of classification has been widely studied \cite{chaudhuri2011differentially, jagannathan2009practical, rubinstein2012learning}. However, these techniques are not suitable when individuals do not trust the data curator completely. To eliminate the need of trusted data curator, techniques to satisfy differential privacy in the local setting have been proposed \cite{bassily2015local, erlingsson2014rappor, kairouz2014extremal, qin2016heavy}. In local differential privacy (LDP), individuals send their data to the data aggregator after privatizing data by perturbation. Hence, these techniques provide plausible deniability for individuals. Data aggregator collects all perturbed values and makes an estimation of statistics such as the frequency of each value in the population. 

In order to guarantee the privacy of the individuals who provide training data in a classification task, we propose using LDP techniques for data collection. We apply LDP techniques to Naive Bayes classifiers which are set of simple probabilistic classifiers based on Bayes' theorem. Naive Bayes classifiers use the assumption of independence between every pair of features. They are highly scalable and particularly suitable when the number of features is high or when the size of training data is small. Naive Bayes is a popular method for text classification (e.g. spam detection and sentiment classification), and it is also used in many other practical applications such as medical diagnosis, digit recognition, and weather prediction. Despite its simplicity, Naive Bayes can often perform better than or close to more sophisticated classification methods.

Given a new instance, Naive Bayes basically computes the conditional probability of each class label, and then assigns the class label with maximum probability to the given instance. Using Bayes' theorem and the assumption of independence of features, each conditional probability can be decomposed as the multiplication of several probabilities. One needs to compute each of these probabilities using training data in order to do Naive Bayes classification. Since the training data must be collected from individuals by preserving privacy, we utilize LDP frequency and statistics estimation methods for collecting perturbed data from individuals and estimating conditional probabilities in Naive Bayes classification. To be able to estimate the conditional probability that a feature would have a specific value given a class label, the relationship between class labels and each feature must be preserved during data collection. Therefore, a new instance can be classified based on the collected privatized training data with Naive Bayes classifier. We developed techniques to perform this privatized training for discrete and continuous data using Naive Bayes classifiers.

Our contributions can be summarized as follows:

First, for the discrete features, we developed LDP Naive Bayes classifier using LDP frequency estimation techniques; where each possible probability that can be used to classify an instance with Naive Bayes is estimated, by preserving the relationships between class labels and features. For perturbation, we utilized five different LDP mechanisms: Direct Encoding (DE), Symmetric and Optimal Unary Encoding (SUE and OUE), Summation with Histogram Encoding (SHE), and Thresholding with Histogram Encoding (THE). 

Second, for the continuous features, we propose two approaches: (a) discretizing the data, and then applying LDP techniques (similar to the previous discussion), and (b) applying Gaussian Naive Bayes after adding Laplace noise to the data to satisfy LDP. For the second approach, we utilized and compared three types of continuous data perturbation methods. In both approaches, we also propose to utilize dimensionality reduction to improve accuracy and to decrease the communication cost and the amount of noise added. 

Third, we conducted experiments with real datasets using various LDP techniques. The results demonstrate that the accuracy of the Naive Bayes classifier is maintained even when the LDP guarantees are satisfied. Our experiment results also show that dimensionality reduction improves classification accuracy without decreasing the privacy level. 


The rest of the paper is organized as follows. We explain Naive Bayes classification, locally differentially private frequency and statistics estimation methods as background in Section \ref{sec:pre}. In Section \ref{sec:method}, we present our methods to apply LDP techniques into Naive Bayes classification. We experimentally evaluate the accuracy of the classification under LDP in Section \ref{sec:eval}. Related work is reviewed in Section \ref{sec:rel}. Finally, Section \ref{sec:conc} concludes the paper.

\section{Preliminaries}
\label{sec:pre}

\subsection{Naive Bayes Classification}
\label{subsec:naive}

In probability theory, Bayes' theorem describes the probability of an event, based on prior knowledge of conditions that might be related to the event. It is stated as follows:
\begin{displaymath}
	P(A~|~B) = \frac{P(B~|~A) \cdot P(A)}{P(B)}
\end{displaymath}
Naive Bayes classification technique uses Bayes' theorem and the assumption of independence between every pair of features. Let the instance to be classified be $n$-dimensional vector $X = \left\{x_1, x_2,..., x_n\right\}$, the names of the features be $F_1, F_2, ..., F_n$, and the possible classes that can be assigned to the instance be $C = \left\{C_1, C_2,..., C_k\right\}$. Naive Bayes classifier assigns the instance $X$ to the class $C_s$ if and only if $P(C_s~|~X) > P(C_j~|~X)$ for $1 \leq j \leq k$ and $j \neq s$. Hence, the classifier needs to compute $P(C_j~|~X)$ for all classes and compare these probabilities. Using Bayes' theorem, the probability $P(C_j~|~X)$ can be calculated as
\begin{displaymath}
	P(C_j~|~X) = \frac{P(X~|~C_j) \cdot P(C_j)}{P(X)}
\end{displaymath}
Since $P(X)$ is same for all classes, it is sufficient to find the class with maximum $P(X~|~C_j) \cdot P(C_j)$. With the assumption of independence of features, it is equal to $P(C_j) \cdot \prod_{i = 1}^{n} P(F_i = x_i~|~C_j)$. Hence, the probability of assigning $C_j$ to given instance is proportional to $P(C_j) \cdot \prod_{i = 1}^{n} P(F_i = x_i~|~C_j)$.

\subsubsection{Discrete Naive Bayes}
\label{subsec:discreteNB}

\begin{table}
   \centering
	\caption{An example dataset}
   \label{tab:ex-db}
	 \begin{tabular}{|c|c|c|c|}
	   \hline
	   \textbf{Age} & \textbf{Income} & \textbf{Gender} & \textbf{Missed Payment} \\
	   \hline
	   \hline
	   Young & Low & Male & Yes \\
	   \hline
   	 Young & High & Female & Yes \\
	   \hline
		 Medium & High & Male & No \\
	   \hline
		 Old & Medium & Male & No \\
		\hline
		 Old & High & Male & No \\
	   \hline
		 Old & Low & Female & Yes \\
	   \hline
		 Medium & Low & Female & No \\
		 \hline
		 Medium & Medium & Male & Yes \\
		 \hline
		 Young & Low & Male & No \\
		 \hline
		 Old & High & Female & No \\
		\hline
	  \end{tabular}
\end{table}

To demonstrate the concept of the naive Bayes classifier for discrete (categorical) data, we use the dataset given in Table \ref{tab:ex-db}. In this example, the classification task is predicting whether a customer will miss a mortgage payment or not. Hence, there are two classes such as $C_1$ and $C_2$ representing missing a previous payment or not, respectively. $P(C_1) = \frac{4}{10}$ and $P(C_2) = \frac{6}{10}$. In addition, conditional probabilities for the feature ``Age'' is given in Table \ref{tablecon}. Similarly, conditional probabilities for the other features can be calculated.

\begin{table}
  \centering
  \caption{Conditional probabilities for $F_1$ (i.e. Age) of the example dataset.}
	\label{tablecon}
  \begin{tabular}{|c|c|} \hline
   $P(Age = Young~|~C_1) = 2/4 $ \\ $P(Age = Young~|~C_2) = 1/6$ \\ \hline
	$P(Age = Medium~|~C_1) = 1/4 $ \\ $P(Age = Medium~|~C_2) = 2/6$ \\ \hline
	$P(Age = Old~|~C_1) = 1/4 $ \\ $P(Age = Old~|~C_2) = 3/6$ \\ \hline 
    \end{tabular}
\end{table}

In order to predict whether a young female with medium income will miss a payment or not, we can set $X = (Age = Young,~Income = Medium,~Gender = Female)$. To use Naive Bayes classifier, we need to compare $P(C_1) \cdot \prod_{i = 1}^{3} P(F_i = x_i~|~C_1)$ and $P(C_2) \cdot \prod_{i = 1}^{3} P(F_i = x_i~|~C_2)$. Since the first one is equal to $0.025$ and the second one is equal to $0.055$, it can be concluded that $C_2$ is assigned for the instance $X$ by Naive Bayes classifier. In other words, it can be predicted that a young female with medium income will not miss her payments.



\subsubsection{Gaussian Naive Bayes}
\label{subsec:gaussNB}
For continuous data, a common approach is assuming the values are distributed according to Gaussian distribution. Then, the conditional probabilities can be computed using the mean and the variance of the values. Let a feature $F_i$ has a continuous domain. For each class $C_j \in C$ the mean $\mu_{i,j}$ and the variance $\sigma^{2}_{i,j}$ of the values of $F_i$ in the training set are computed. For the given instance $X$, the conditional probability $P(F_i = x_i~|~C_j)$ is computed using Gaussian distribution as follows:
\begin{displaymath}
	P(F_i = x_i~|~C_j) = \frac{1}{\sqrt{2\pi\sigma^{2}_{i,j}}} e^{-\frac{(x_i-\mu_{i,j})^2}{2\sigma^{2}_{i,j}}}
\end{displaymath}
Gaussian Naive Bayes can also be used for features with large discrete domain. Otherwise, the accuracy may reduce because of the high number of values which are not seen in the training set.

\subsection{Local Differential Privacy}
\label{subsec:ldp}

Local differential privacy (LDP) is a way of measuring the individual privacy in the case where the data curator is not trusted. In LDP setting, individuals perturb their data before sending it to a data aggregator. Hence, the data aggregator only sees perturbed data. It aggregates all reported values and \textit{estimates} privacy-preserving statistics. LDP states that for any reported value, the probability of distinguishing two input values by the data aggregator is at most $e^{-\epsilon}$. The formal definition of local differential privacy is as follows:

\begin{mydef}
A protocol $P$ satisfies $\epsilon$-local differential privacy if for any two input values $v_1$ and $v_2$ and any output $o$ in the output space of $P$,
\begin{displaymath}
\mathrm{Pr}\left[P(v_1) = o\right] \leq \mathrm{Pr}\left[P(v_2) = o\right] \cdot e^{\epsilon}
\end{displaymath}
\end{mydef}

Randomized response mechanism is one method to satisfy LDP. In the binary randomized response mechanism, the input is a single bit. An individual sends the correct bit to the data aggregator with probability $p$ and incorrect bit with probability $1-p$. The aggregator can estimate the actual number of 0s and 1s by using the probability $p$ and the reported numbers of 0s and 1s. To satisfy $\epsilon$-LDP, $p$ can be selected as $\frac{e^\epsilon}{1 + e^\epsilon}$. This problem can be generalized into frequency estimation problem where the inputs can be selected from a larger set containing more than two values.

\subsubsection{LDP Frequency Estimation}
\label{subsubsec:ldpFreq}

In the problem of frequency estimation, there are $m$ individuals having a value from the set $\mathcal{D} = \left\{1,2,...,d\right\}$. The aim of data aggregator is to find the number of individuals having a value $i \in \mathcal{D}$ for all values in the set. Wang et al. \cite{wang2017locally} proposed a framework to generalize the LDP frequency estimation protocols in the literature, and they also proposed two new protocols. Here, we summarize the LDP protocols which are explained in \cite{wang2017locally} in detail. All of them can be used for frequency estimation in our solution. We empirically compare their effect on accuracy in our problem setting in Section \ref{sec:eval}.

\textbf{Direct encoding (DE):} In this method, there is no encoding of input values. For perturbation, an individual reports her value $v$ correctly with probability $p = \frac{e^\epsilon}{e^\epsilon + d - 1}$, or reports one of the remaining $d-1$ values with probability $q = \frac{1}{e^\epsilon + d - 1}$ per each. When the aggregator collects all perturbed values from $m$ individuals, it estimates the frequency of each $i \in \left\{1,2,...,d\right\}$ as follows: Let $c_i$ be the number of times $i$ is reported. Estimated number of occurrence of value $i$ in the population is computed as $E_i = \frac{c_i - m \cdot q}{p - q}$.

\textbf{Histogram encoding:} An individual encodes her value $v$ as length-$d$ vector $\left[0.0,....,1.0,...,0.0\right]$ where only $v^{\mathrm{th}}$ component is $1.0$ and the remaining are $0.0$. Then, she perturbs her value by adding Lap($\frac{2}{\epsilon}$) to each component in the encoded value, where Lap($\frac{2}{\epsilon}$) is a sample from Laplace distribution with mean 0 and scale parameter $\frac{2}{\epsilon}$. When the data aggregator collects all perturbed values, it can use two estimation methods. In summation with histogram encoding (SHE), it calculates the sum of all values reported by individuals. To estimate the number of occurrence of value $i$ in the population, the data aggregator sums the $i^{\mathrm{th}}$ components of all reported values. In thresholding with histogram encoding (THE), the data aggregator sets all values greater than a threshold $\theta$ to 1, and the remaining to 0. Then it estimates the number of $i$'s in the population as $E_i = \frac{c_i - m \cdot q}{p - q}$, where $p = 1 - \frac{1}{2} e^{\frac{\epsilon}{2}(1 - \theta)}$, $q = \frac{1}{2} e^{-\frac{\epsilon}{2}\theta}$, and $c_i$ is the number of $1$'s in the $i^{\mathrm{th}}$ components of all reported values after applying thresholding. 

\textbf{Unary encoding:} In this method, an individual encodes her value $v$ as length-$d$ binary vector $\left[0,....,1,...,0\right]$ where only $v^{\mathrm{th}}$ bit is $1$ and the remaining are $0$. Then, for each bit in the encoded vector, she reports correctly with probability $p$ and incorrectly with probability $1-p$ if the input bit is $1$. Otherwise, she reports correctly with probability $1-q$ and incorrectly with probability $q$. In symmetric unary encoding (SUE), $p$ is selected as $\frac{e^{\epsilon / 2}}{e^{\epsilon / 2} + 1}$ and $q$ is selected as $1-p$. In optimal unary encoding (OUE), $p$ is selected as $\frac{1}{2}$ and $q$ is selected as $\frac{1}{e^{\epsilon} + 1}$. The data aggregator estimates the number of $1$'s in the population as $E_i = \frac{c_i - m \cdot q}{p - q}$, where $c_i$ denotes the number of $1$'s in the $i^{\mathrm{th}}$ bit of all reported values.

\subsubsection{LDP Mean Estimation}
\label{subsubsec:ldpMean}

As explained in Section \ref{subsec:gaussNB}, Gaussian Naive Bayes is suitable for large discrete domains and continuous domains. Conditional probabilities are computed using the mean and the variance. In order to compute the mean under LDP, Laplace mechanism can be used \cite{nguyen2016collecting}. Let the domain be normalized, and an individual has a value $v \in \left[-1,1\right]$. The individual adds Laplace noise Lap($\frac{2}{\epsilon}$) to her value and reports noisy value ($v' = v + \textrm{Lap}(\frac{2}{\epsilon}$)) to the data aggregator. Since the mean of noises that are drawn from Laplace distribution is 0, the data aggregator calculates the sum of all noisy values reported by individuals, and divides the sum by the number of individuals to estimate the mean. As for estimating the variance, we explain our proposed method in Section \ref{subsec:conNB}.

\subsubsection{LDP with Multi-dimensional Data}
\label{subsubsec:ldpMulti}

The frequency and mean estimation methods described in Section \ref{subsubsec:ldpFreq} and \ref{subsubsec:ldpMean} work for one-dimensional data. If the data owned by individuals is multi-dimensional, reporting each value with these methods may cause privacy leaks due to the dependence of features. Hence, the following approaches were proposed to deal with $n$-dimensional data.



\textbf{Approach 1:} For the Laplace mechanism described in Section \ref{subsubsec:ldpMean}, LDP can also be satisfied if the noise scaled with the number of dimensions $n$ \cite{nguyen2016collecting}. Hence, if an individuals' input is $V = \left(v_1,...,v_n\right)$ such that $v_i \in \left[-1,1\right]$ for all $i \in \left\{1,...,n\right\}$, then she can report each $v_i$ after adding $\textrm{Lap}(\frac{2n}{\epsilon})$ (i.e. $v_i' = v_i + \textrm{Lap}(\frac{2n}{\epsilon}$)). This approach is not suitable if the number of dimensions $n$ is high because large amount of noise reduces the accuracy. 

\textbf{Approach 2:} For mean estimation, Nguy{\^e}n et al. \cite{nguyen2016collecting} introduced an algorithm that requires reporting one bit by each individual to the data aggregator. An individual has an input value $V = \left(v_1,...,v_n\right)$ such that $v_i \in \left[-1,1\right]$ for all $i \in \left\{1,...,n\right\}$. She can perturb and report her input as follows:
\begin{itemize}
	\item She select $j \in \left\{1,...,n\right\}$ uniformly at random.
	\item She samples Bernoulli variable $u$ such that $\textrm{Pr}\left[u=1\right] = \frac{v_j \left(e^{\epsilon} - 1\right) + e^{\epsilon} + 1}{2e^{\epsilon} + 2}$.
	\item She sets $v_j' = \frac{e^{\epsilon} + 1}{e^{\epsilon} - 1} \cdot n$ if $u = 1$, $v_j' = - \frac{e^{\epsilon} + 1}{e^{\epsilon} - 1} \cdot n$ otherwise.
	\item She reports $V' = \left(0,...,0,v_j',0,...,0\right)$ to the data aggregator.
\end{itemize}
Since the only non-zero value is $v_j'$ and it has two possible values, it is sufficient to report one bit to indicate the sign of $v_j'$. Each feature is approximately reported by $\frac{m}{n}$ individuals. This approach is efficient in terms of communication cost.

\textbf{Approach 3:} 
The first two approaches are specific to continuous data. Hence, we outline a third approach that is more general. The data aggregator requests only one perturbed input from each individual to satisfy $\epsilon$-LDP. Each individual can select the input to be reported uniformly at random or the data aggregator can divide the individuals into $n$ groups and requests different input values from each group. As a result, each feature is approximately reported by $\frac{m}{n}$ individuals. This approach is suitable when the number of individuals $m$ is high relative to the number of features $n$. Otherwise the accuracy decreases since the number of reported values is low for each feature.

\subsection{Dimensionality Reduction}
\label{subsec:dr}

The approaches for dealing with multi-dimensional data suffer from the high number of dimensions which necessitates adding more noise that results in decreasing the accuracy. In the first approach, the amount of noise is directly proportional to the number of dimensions. In the second approach, the number of individuals who report each feature decreases for high number of dimensions because each feature is approximately reported by $\frac{m}{n}$ individuals. Therefore, we propose to utilize dimensionality reduction techniques to improve accuracy. Dimensionality reduction is a machine learning tool that is traditionally used to solve over-fitting issues, and to reduce the computational cost caused by high numbers of features. We utilize two commonly used methods for dimensionality reduction: Principal Component Analysis (PCA) and Discriminant Component Analysis (DCA) \cite{kung2014kernel}.

\textbf{PCA} reduces the dimensions while preserving most of the information by projecting the data on the principal components with the highest variance. By projecting the data in the direction of the highest variability, PCA also tends to decrease the reconstruction error; thus improving recoverability of the original data from its projection. On the other hand, \textbf{DCA} utilizes the class labels $C_i$'s to project the data in the direction that can effectively discriminate between different classes. Such direction might not be necessarily the direction of the highest variance; thus DCA can be superior to PCA for labeled data.

\section{Naive Bayes Classification under Local Differential Privacy}
\label{sec:method}

As explained in Section \ref{subsec:naive}, one needs to know the probability $P(C_j)$ for all classes, and $P(F_i = x_i~|~C_j)$ for all classes and all possible $x_i$ values in order to use Naive Bayes classifier. These probabilities are calculated based on the training data. However, when individuals avoid sharing their data for training due to privacy reasons, it is impossible to calculate these probabilities. Since LDP provides plausible deniability for individuals, LDP methods can be used to train Naive Bayes classifier. In this section, we explain the estimation of such necessary probabilities using LDP methods. First we introduce a solution for classification for all discrete features (Section \ref{subsec:disNB}), and then we explain the solutions to deal with continuous data (Section \ref{subsec:conNB}). Table \ref{tablenot} shows the notations used in the paper. 

\begin{table}
  \centering
  \caption{Notations used in the paper.}
	\label{tablenot}
  \begin{tabular}{|c|l|} \hline
   $X = (x_1,...,x_n)$ & instance to be classified \\ \hline
	$C = \left\{C_1, C_2,..., C_k\right\}$ & the set of class labels \\ \hline
	$n$ & the number of features \\ \hline
	$k$ & the number of class labels \\ \hline
	$m$ & the number of individuals \\ \hline
    \end{tabular}
\end{table}

\subsection{LDP Naive Bayes with Discrete Features}
\label{subsec:disNB}

We initially consider the case where all the features are numerical and discrete. There are $m$ individuals who are reluctant to share their data to train a classifier. However, they can share perturbed data to preserve their privacy. By satisfying LDP during data collection, the privacy of individuals can be guaranteed. Here, we propose a solution that utilizes the LDP frequency estimation methods given in Section \ref{subsec:ldp} in order to compute all necessary probabilities for a Naive Bayes classifier.

The data aggregator needs to estimate class probabilities $P(C_j)$ for all classes in $C = \left\{C_1, C_2,..., C_k\right\}$ and conditional probabilities $P(F_i = x_i~|~C_j)$ for all classes and all possible $x_i$ values. Let an individual's (e.g. Alice's) data be $(a_1, a_2,..., a_n)$ and her class label be $C_v$. She needs to prepare her input and perturb it by satisfying LDP. We now explain the preparation and the perturbation of input values based on Alice's data and the estimation of the class probabilities and the conditional probabilities by data aggregator.

\subsubsection{Computation of Class Probabilities}
\label{subsubsec:classprob}

For the computation of class probabilities, Alice's input becomes $v \in \left\{1,2,...,k\right\}$ since her class label is $C_v$. Alice encodes and perturbs her value $v$, and reports to the data aggregator. Any LDP frequency estimation method which is explained in Section \ref{subsubsec:ldpFreq} can be used. Similarly, other individuals report their perturbed class labels to the data aggregator. The data aggregator collects all perturbed data and estimates the frequency of each value $j \in \left\{1,2,...,k\right\}$ as $E_j$. As a result, the probability $P(C_j)$ is estimated as $\frac{E_j}{\sum_{i=1}^{k} E_i}$. For the example dataset in Table \ref{tab:ex-db}, Alice's input $v$ becomes $1$ if she has a missing payment or $2$ if she does not have a missing payment.


\subsubsection{Computation of Conditional Probabilities}
\label{subsubsec:condprob}

To estimate the conditional probabilities $P(F_i = x_i~|~C_j)$, it is not sufficient to report feature values directly. To be able to compute these probabilities, the relationship between class labels and features must be preserved. To keep this relationship, individuals prepare their inputs using feature values and class labels. Let the total number of possible values for $F_i$ be $n_i$. If Alice's value in $i^{\mathrm{th}}$ dimension is $a_i \in \left\{1,2,...,n_i\right\}$ and her class label value is $v \in \left\{1,2,...,k\right\}$, then Alice's input for feature $F_i$ becomes $v_i = (a_i - 1) \cdot k + v$. Therefore, each individual calculates her input for the $i^{\mathrm{th}}$ feature in the range of $\left[1,k \cdot n_i\right]$. For instance, let ``Age'' values in the Table \ref{tab:ex-db} be enumerated as (Young = $1$), (Medium = $2$), (Old = $3$). For this feature, an individual's input can be a value between 1 and 6, where 1 represents the age is young and there is a missing payment, and 6 represents the age is old and there is no missing payment. Therefore, there is one input value that corresponds to each line of Table \ref{tablecon}. Similarly, the number of possible inputs for ``Income'' is 6 and the number of possible inputs for ``Gender'' is 4. After determining her input in $i^{\mathrm{th}}$ feature, Alice encodes and perturbs her value $v_i$, and reports the perturbed value to the data aggregator. To estimate the conditional probabilities for $F_i$, the data aggregator estimates the frequency of individuals having value $y \in \left\{1,2,...,n_i\right\}$ and class label $z \in \left\{1,2,...,k\right\}$ as $E_{y,z}$ by estimating the frequency of input $(y-1) \cdot k + z$. Hence, the conditional probability $P(F_i = x_i~|~C_j)$ is estimated as $\frac{E_{x_i,j}}{\sum_{h=1}^{n_i} E_{h,j}}$. For the example given above, to estimate the probability $P(Age = Medium~|~C_2)$, the data aggregator estimates the frequency of 2, 4, and 6 as $E_{1,2}$, $E_{2,2}$, and $E_{3,2}$, respectively. Then $P(Age = Medium~|~C_2)$ is estimated as $\frac{E_{2,2}}{E_{1,2} + E_{2,2} + E_{3,2}}$. 

As a result, in order to contribute to the computation of class probabilities and conditional probabilities, each individual can prepare $n + 1$ inputs (i.e. $\left\{v,~v_1,~v_2,~....,~v_n\right\}$ for Alice) that can be reported after perturbation. As mentioned in Section \ref{subsubsec:ldpMulti}, reporting multiple values which are dependent to each other decreases the privacy level. Reporting all $n+1$ perturbed values increases the probability of predicting the class labels of individuals by the data aggregator. This case is similar to requesting multiple queries in the centralized setting of differential privacy. Hence, each individual reports one input as described in Approach 3 in Section \ref{subsubsec:ldpMulti}. 

Finally, when the data aggregator estimates a value such as $E_j$ or $E_{y,z}$, the estimation may give a negative result. In that case, we set all the negative estimations to $1$ to obtain valid probability.

\subsection{LDP Naive Bayes with Continuous Features}
\label{subsec:conNB}

\begin{figure*}%
\centering
\includegraphics[width=\textwidth]{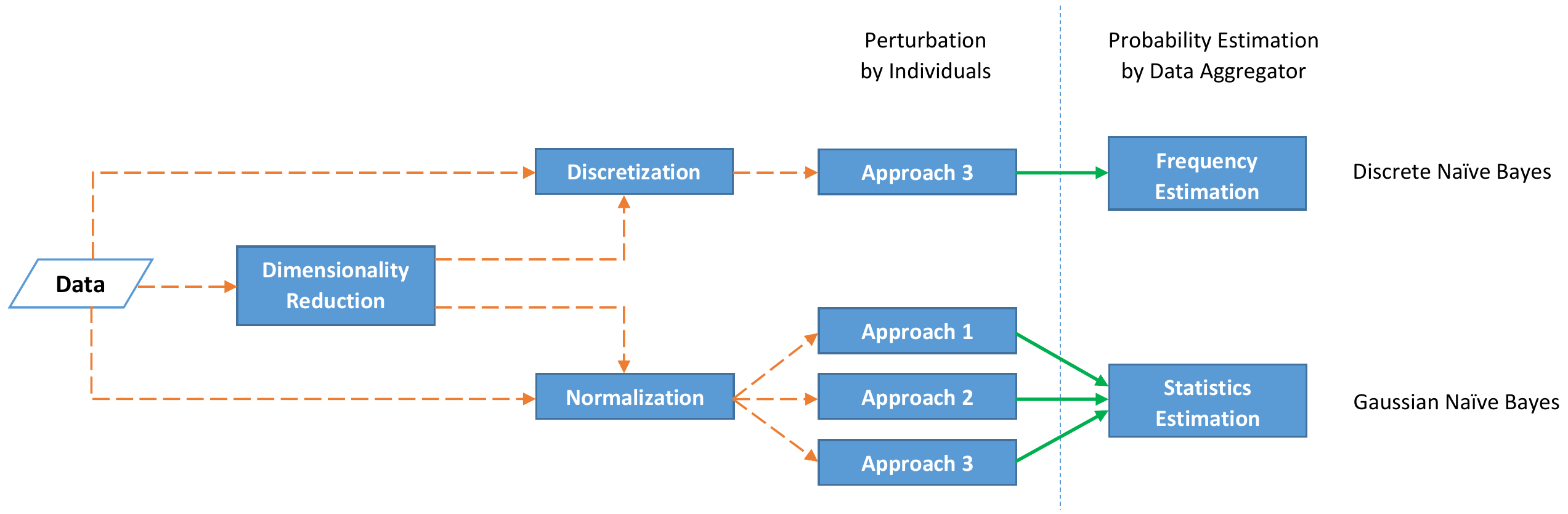}%
\caption{Steps of LDP Naive Bayes for multi-dimensional continuous data.}%
\label{fig:steps}%
\end{figure*}

In order to satisfy LDP in Naive Bayes classification for continuous data, we propose two different solutions. First solution is discretizing the continuous data and applying the discrete Naive Bayes solution outlined in Section \ref{subsec:disNB}. In this solution, continuous numerical data is divided into buckets to make it finite and discrete. Each individual perturbs her input after discretization. Second, the data aggregator can use Gaussian Naive Bayes to estimate the probabilities as given in Section \ref{subsec:gaussNB}. To estimate the mean and the variance, the data aggregator uses LDP methods given in Section \ref{subsubsec:ldpMean}. Figure \ref{fig:steps} shows the steps of the proposed solutions. As explained in Section \ref{subsubsec:ldpMulti}, the number of dimensions can be reduced to improve accuracy; hence, we utilize dimensionality reduction techniques. Now, we describe the solutions in detail.

\textbf{Discrete Naive Bayes.} We first propose to use the solution introduced for discrete data in Section \ref{subsec:disNB}. Based on known feature ranges for features with continuous or large domain, the data aggregator determines the intervals for buckets in order to discretize the domain. Equal-Width Discretization (EWD) can be used for equally partitioning the domain. EWD computes the width of each bin as $\frac{max - min}{n_b}$ where $max$ and $min$ are the maximum and minimum feature values, and $n_b$ is the number of desired bins. We utilized EWD in our experiments for discretization.

When the data aggregator shares the intervals with individuals, each individual firstly discretizes her continuous feature values, and then applies the procedure described in Section \ref{subsec:disNB} for perturbation. The data aggregator also estimates the probabilities with the same procedure for LDP Naive Bayes for discrete data. As mentioned in Section \ref{subsubsec:condprob}, each individual should report just one perturbed value to guarantee $\epsilon$-LDP.

\textbf{Gaussian Naive Bayes.} As explained in Section \ref{subsec:gaussNB}, a common approach for Naive Bayes classification for continuous data is assuming the data is normally distributed. For locally differentially private Gaussian Naive Bayes, computing the class probabilities is same with the computation for discrete features as given in Section \ref{subsubsec:classprob}. To compute conditional probabilities, the data aggregator needs to have the mean and the variance of training values for each feature given a class label. That is, to compute 
$P(F_i = x_i~|~C_j)$, the data aggregator needs to estimate the mean $\mu_{i,j}$ and the variance $\sigma^{2}_{i,j}$ using the $F_i$ values of individuals with a class label $C_j$. Hence, the association between features and class labels has to be maintained (similar to the discrete Naive Bayes classifier).

The mean estimation was explained in Section \ref{subsubsec:ldpMean}. However to compute the mean $\mu_{i,j}$ and the variance $\sigma^{2}_{i,j}$ together, we propose the following method: the data aggregator divides the individuals into two groups. One group contributes to the estimation of the mean (i.e. $\mu_{i,j}$) by perturbing their inputs and sharing with the data aggregator, while the other group contributes to the estimation of the mean of squares (i.e. $\mu^s_{i,j}$) by perturbing the squares of their inputs and sharing with data aggregator. 

Let Bob has class label $C_j$ and his feature $F_i$ value be $b_i$. Note that, the domain of each feature was assumed to be normalized to have a value in $\left[-1,1\right]$. If Bob is in the first group, he adds Laplace noise to his value $b_i$ and obtains perturbed feature value $b_i'$. When data aggregator collects all perturbed feature values from individuals in the first group having class label $C_j$, it computes the mean of the perturbed feature values which gives an estimation of the mean $\mu_{i,j}$ because the mean of noise added by individuals is 0. Similar operations could be followed by the second group. If Bob is in the second group, he adds noise to his squared value $b_i^2$ to obtain ${b_i^2}'$ and shares it with the data aggregator. Similarly, the data aggregator computes the estimation of the  mean of squares ($\mu^s_{i,j}$). Finally, the variance $\sigma^{2}_{i,j}$ can be computed as $\mu^s_{i,j} -(\mu_{i,j})^2$. Once again, each individual reports only one of her value or square of her value after perturbation because they are dependent values.

In this explained method to compute the mean and the variance, the class label of individuals are not hidden from the data aggregator. To hide the class labels, we adopt the following approach: an individual (Bob) reporting a feature value $F_i = b_i$ associated with class $C_j$ where $j \in \left\{ 1, 2, \cdots, k\right\}$, first constructs a vector of length $k$ where $k$ is the number of class labels. The vector is initialized to zeros except for the $j^{\mathrm{th}}$ element corresponding to the $j^{\mathrm{th}}$ class label which is set to the feature value $b_i$. After that, each element of the vector is perturbed as usual (i.e. by adding Laplace noise), and contributed to the data aggregator. Since noise is added even to the zero elements of the vector, the data aggregator will not be able to deduce the actual class label, or the actual values.

As for estimating the actual mean value (and mean of the squared values) for each class, the data aggregator only needs to compute the mean of the perturbed values as usual, and then dividing that value by the probability of that class. To understand why, assume that a specific class $j$ has Probability $P(C_j)$ (explained in Section \ref{subsubsec:classprob}). Hence, for a specific feature $F_i$, only $P(C_j)$ of the individuals have their actual values in $j^{\mathrm{th}}$ element of the input vector, while the remaining proportion ($1 - P(C_j)$) have zeros. Hence, after the noise clustered around the actual mean cancels each other, and the noise clustered around zero cancel each other, we would have $P(C_j) * \mu_{i,j} = observed ~(shifted)~mean$. Hence, we can divide the observed mean by $P(C_j)$ to obtain the estimated mean. The same situation applies for the mean of the squared values, and hence for computing the variance.

\section{Experimental Evaluation}
\label{sec:eval}

To evaluate the accuracy of Naive Bayes classification under local differential privacy, we have implemented the proposed methods in Python utilizing pandas and NumPy libraries. We have implemented $5$ different LDP protocols for frequency estimation such as Direct Encoding (DE), Summation with Histogram Encoding (SHE), Thresholding with Histogram Encoding (THE), Symmetric Unary Encoding (SUE), and Optimal Unary Encoding (OUE) which are presented in Section \ref{subsec:ldp}. We performed experiments with different $\theta$ values in THE and we achieved best accuracy when $\theta = 0.25$. Hence, we give the experiment results of SHE for $\theta = 0.25$. We repeated all experiments 100 times and present the average classification accuracy. We used datasets from UCI Machine Learning repository \cite{Dua:2017} and selected $80\%$ of the datasets for training and the remaining $20\%$ for testing. We firstly present the results for the datasets with categorical and discrete features in Section \ref{sec:evalDiscrete}. The results for continuous data is given in Section \ref{sec:evalContinuous}.

\subsection{LDP Naive Bayes with Discrete Features}
\label{sec:evalDiscrete}

To evaluate the classification accuracy of the proposed method in Section \ref{subsec:disNB} for classifying data with discrete features, we used Car Evaluation, Chess, Mushroom, and Connect-4 datasets from UCI ML repository. The number of instances, features, and class labels are given in Table \ref{table:dataset}. Initially, we performed Naive Bayes classification without local differential privacy to compare the accuracy under local differential privacy. 

\begin{table}
  \centering
  \caption{Datasets used in the experiments.}
	\label{table:dataset}
  \begin{tabular}{|c|c|c|c|} \hline
  Name & \# Instances & \# Features & \# Class Labels\\ \hline \hline
	Car Evaluation & $1,728$ & $6$ & $4$ \\ \hline
	Chess & $3,196$ & $36$ & $2$ \\ \hline
	Mushroom & $8,124$ & $22$ & $2$ \\ \hline
	Connect-4 & $67,557$ & $42$ & $3$ \\ \hline
    \end{tabular}
\end{table}

\begin{figure*}[!ht]
\begin{subfigure}{.5\textwidth}
  \centering
  \includegraphics[trim=0 0 0 0,clip,width=7cm]{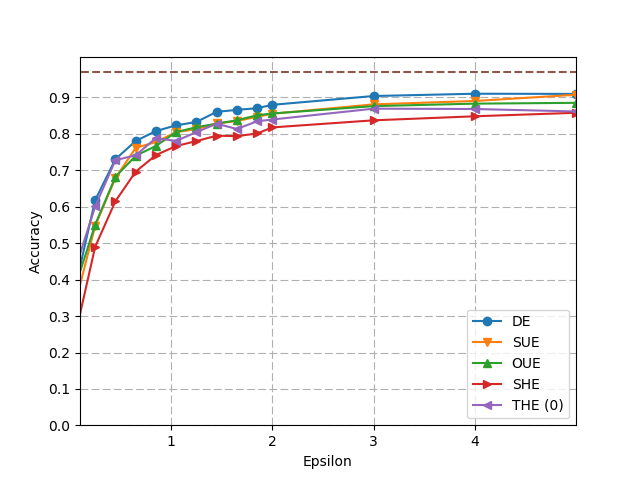}
  \caption{Car Evaluation dataset}
  \label{car1}
\end{subfigure}%
\begin{subfigure}{.5\textwidth}
  \centering
  \includegraphics[trim=0 0 0 0,clip,width=7cm]{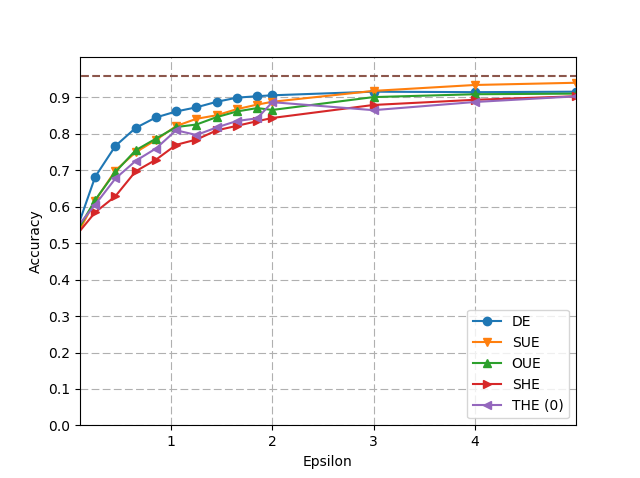}
  \caption{Chess dataset}
  \label{chess1}
\end{subfigure}%

\begin{subfigure}{.5\textwidth}
  \centering
  \includegraphics[trim=0 0 0 0,clip,width=7cm]{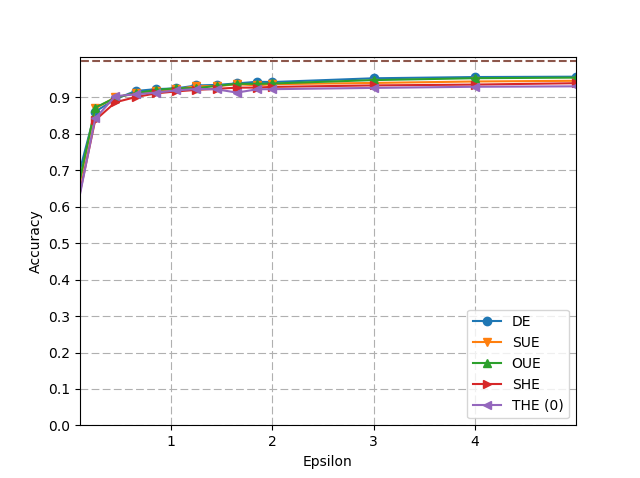}
  \caption{Mushroom dataset}
  \label{mushroom1}
\end{subfigure}%
\begin{subfigure}{.5\textwidth}
  \centering
  \includegraphics[trim=0 0 0 0,clip,width=7cm]{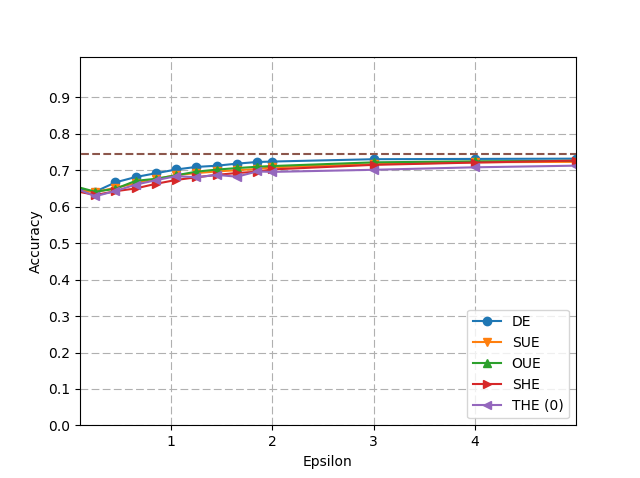}
  \caption{Connect-4 dataset}
  \label{connect1}
\end{subfigure}%
\vspace{-3mm}
\caption{Classification accuracy for datasets with discrete features}
\label{fig:ldp-discrete}
\vspace{-3mm}
\end{figure*}

%
%
%

Experiment results for varying $\epsilon$ values up to 5 are shown in Figure \ref{fig:ldp-discrete}. Dotted lines in the figures show the accuracy without privacy. As expected, when the number of instances in the training set increases, the accuracy is better for smaller $\epsilon$ values. For instance, in Connect-4 dataset, all protocols except SHE provide more than $65\%$ accuracy even for very small $\epsilon$ values. Since the accuracy without privacy is approximately $75\%$, the accuracy of all of these protocols for $\epsilon$ values smaller than $1$ is noticeable. The results are also similar for Mushroom dataset. For $\epsilon = 0.5$, all protocols except SHE provide nearly $90\%$ classification accuracy. In all of the datasets, the protocol with worst accuracy is SHE. Since this protocol simply sums the all noisy values, its variance is higher than the other protocols. DE achieves the best accuracy for small $\epsilon$ values in Car Evaluation and Chess datasets because the input domains are small. The variance of DE is proportional to the size of the input domain. Therefore, its accuracy is better when the input domain is small. SUE and OUE provides similar accuracy in all of the experiments. They perform better than DE when the size of input domain is large. Although OUE is proposed by \cite{wang2017locally} to decrease variance, we did not observe considerable utility difference between SUE and OUE in our experiments.

\subsection{LDP Naive Bayes with Continuous Features}
\label{sec:evalContinuous}

\begin{figure*}[!ht]
\begin{subfigure}{.5\textwidth}
  \centering
  \includegraphics[trim=0 0 0 0,clip,width=7cm]{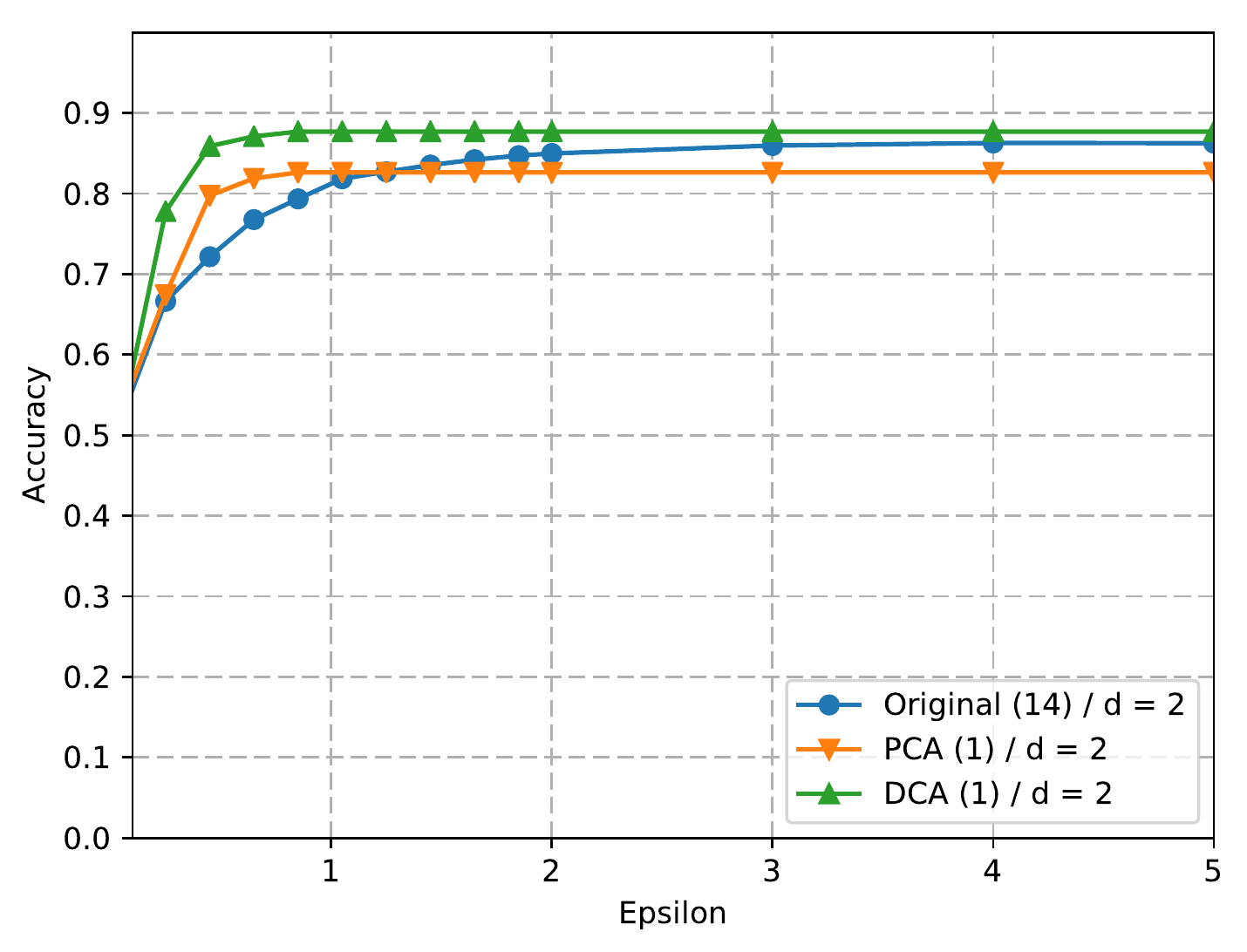}
  \caption{Australian dataset / Direct Encoding}
  \label{ausDE}
\end{subfigure}%
\begin{subfigure}{.5\textwidth}
  \centering
  \includegraphics[trim=0 0 0 0,clip,width=7cm]{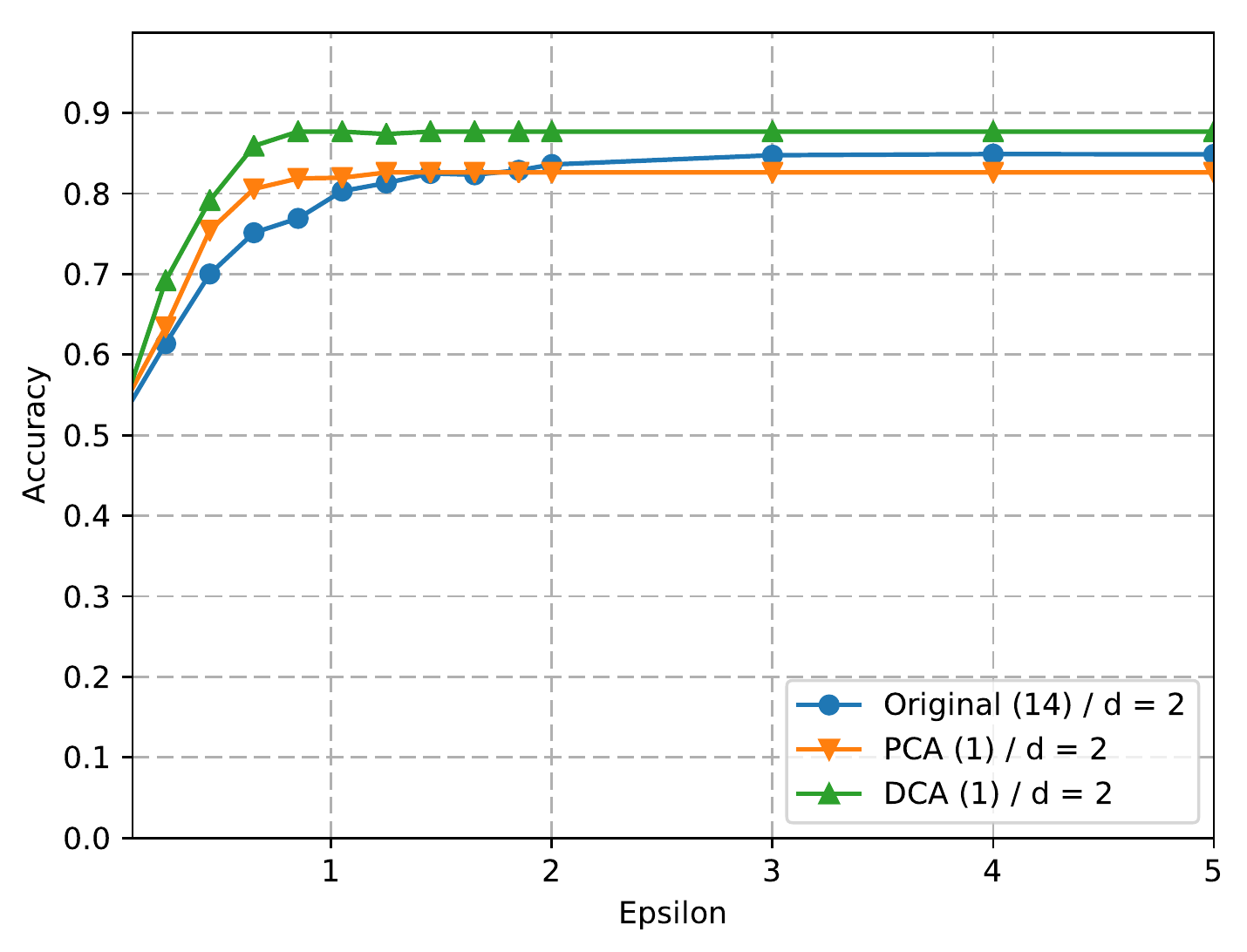}
  \caption{Australian dataset / Optimized Unary Encoding}
  \label{ausOUE}
\end{subfigure}%

\begin{subfigure}{.5\textwidth}
  \centering
  \includegraphics[trim=0 0 0 0,clip,width=7cm]{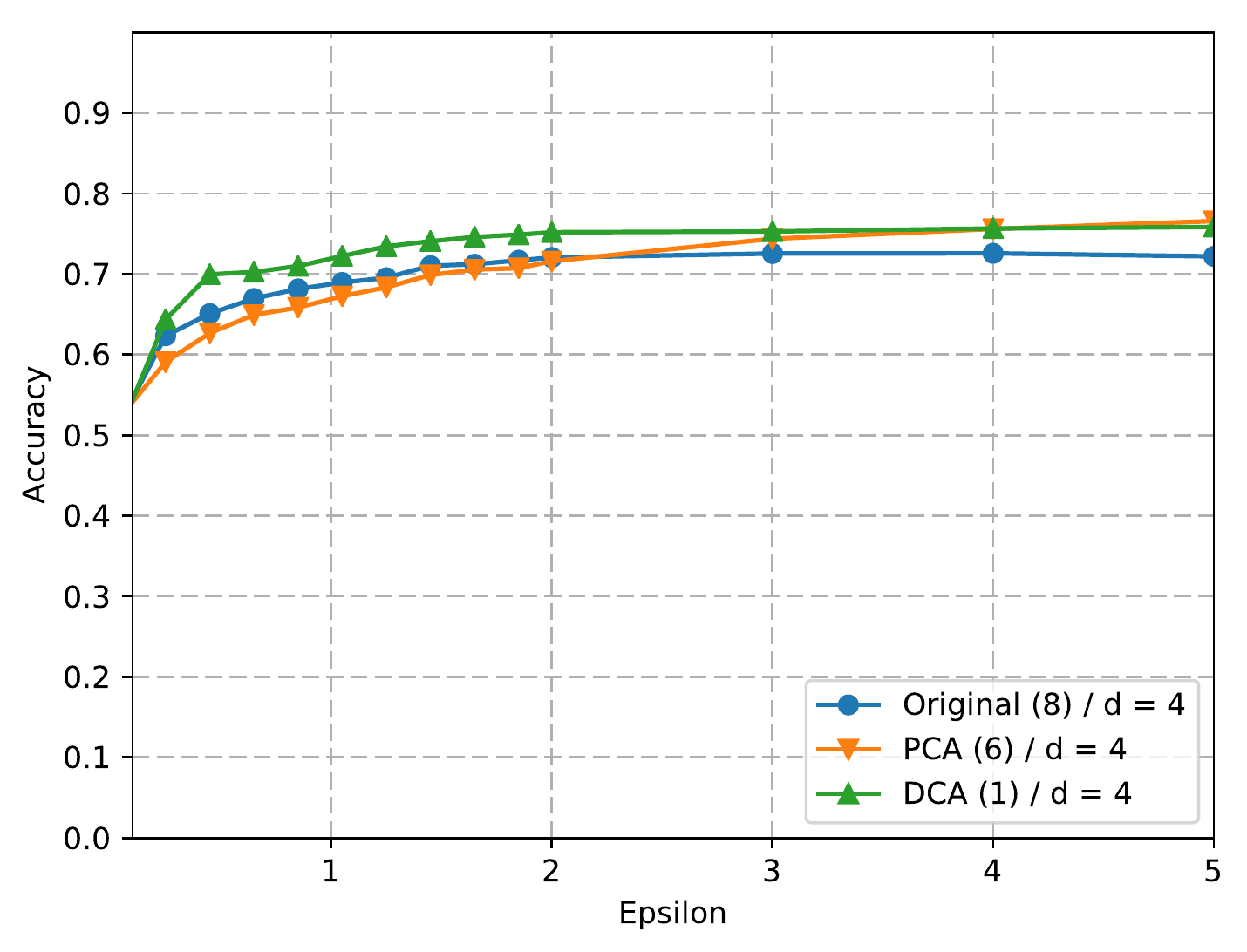}
  \caption{Diabetes dataset / Direct Encoding}
  \label{diabDE}
\end{subfigure}%
\begin{subfigure}{.5\textwidth}
  \centering
  \includegraphics[trim=0 0 0 0,clip,width=7cm]{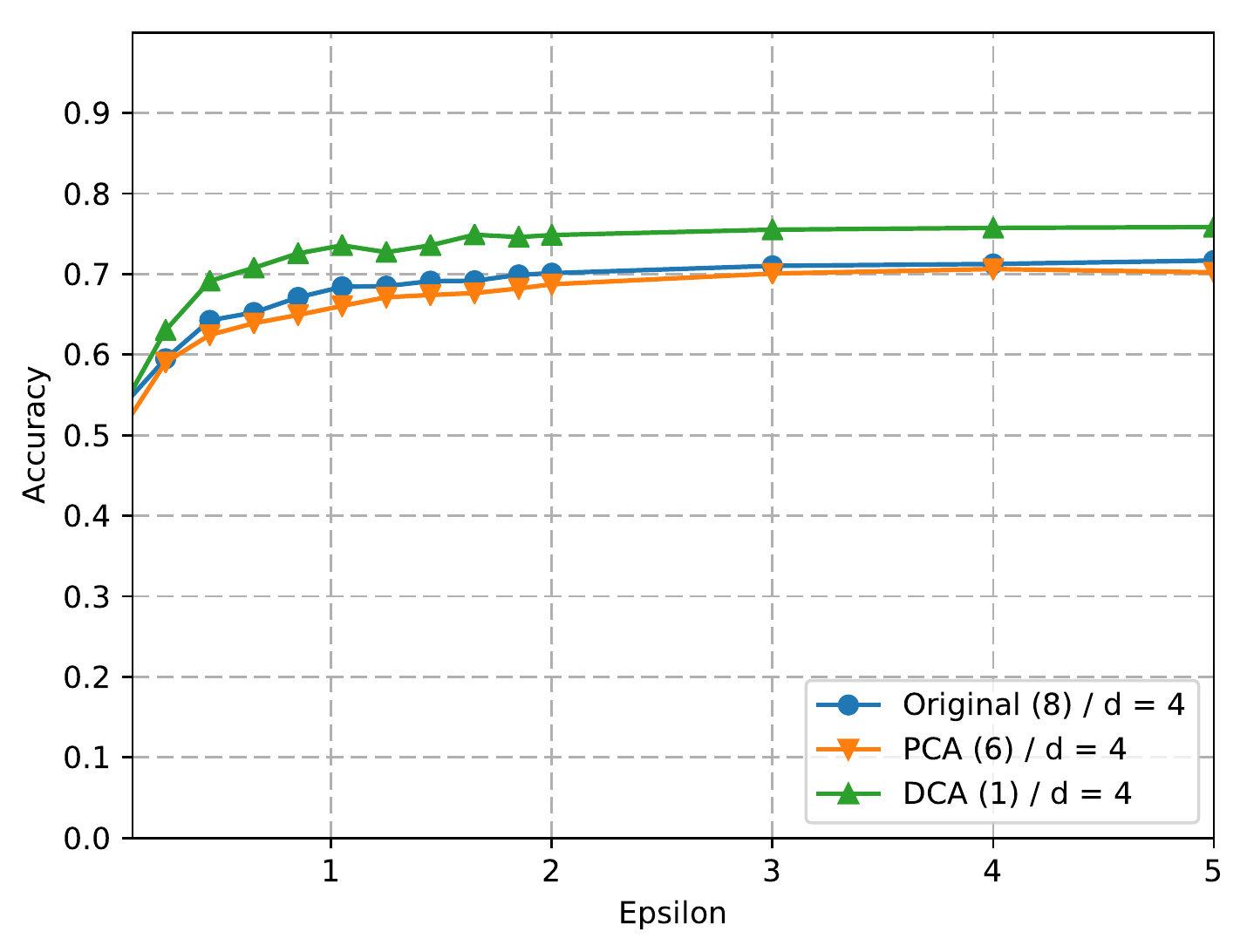}
  \caption{Diabetes dataset / Optimized Unary Encoding}
  \label{diabOUE}
\end{subfigure}%
\vspace{-3mm}
\caption{Classification accuracy for datasets with continuous features using discretization}
\vspace{-3mm}
\label{fig:ldp-cont-discretized}
\end{figure*}

In this section, we outline the results for the methods proposed in Section \ref{subsec:conNB} for continuous data. We conducted the experiments on two different datasets: Australian and Diabetes. The Australian dataset has 14 original features, and the Diabetes dataset has 8 features. Initially, we applied the discretization method and implemented two dimensionality reduction techniques (i.e. PCA and DCA) to observe the effect of them in accuracy. The results for two datasets for different values of $\epsilon$ are given in Figure \ref{fig:ldp-cont-discretized}. We present the results for two LDP schemes (i.e. Direct Encoding and Optimized Unary Encoding) which provide the best accuracy for different domain sizes. ,The input domain is divided into $d=2$ buckets for Australian dataset and $d=4$ buckets for Diabetes dataset. For Australian dataset, we obtained the best results for PCA and DCA when the number of features is reduced to one. For Diabetes dataset, best accuracy is achieved when PCA reduces the number of features to 6 and when DCA reduces the number of features to one. As evident in Figure \ref{fig:ldp-cont-discretized}, DCA provides the best classification accuracy, which shows the advantage of using dimensionality reduction before discretization. As expected, DCA's accuracy is better than PCA since it is mainly designed for classification.

\begin{figure*}[!ht]
\begin{subfigure}{.33\textwidth}
  \centering
  \includegraphics[trim=0 0 0 0,clip,width=\linewidth]{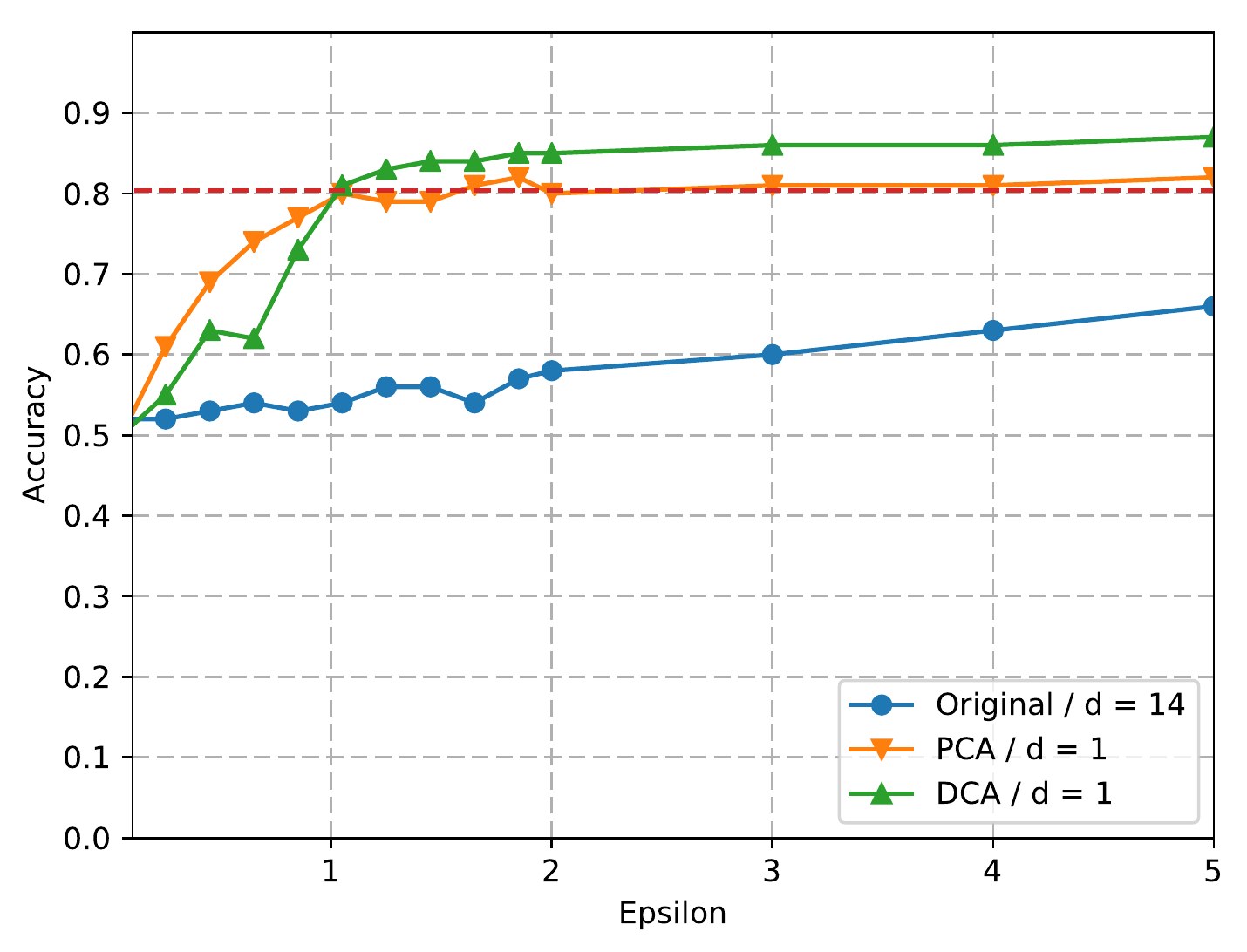}
  \caption{Australian dataset / Approach 1}
  \label{fig:australian_basicOne}
\end{subfigure}%
\begin{subfigure}{.33\textwidth}
  \centering
  \includegraphics[trim=0 0 0 0,clip,width=\linewidth]{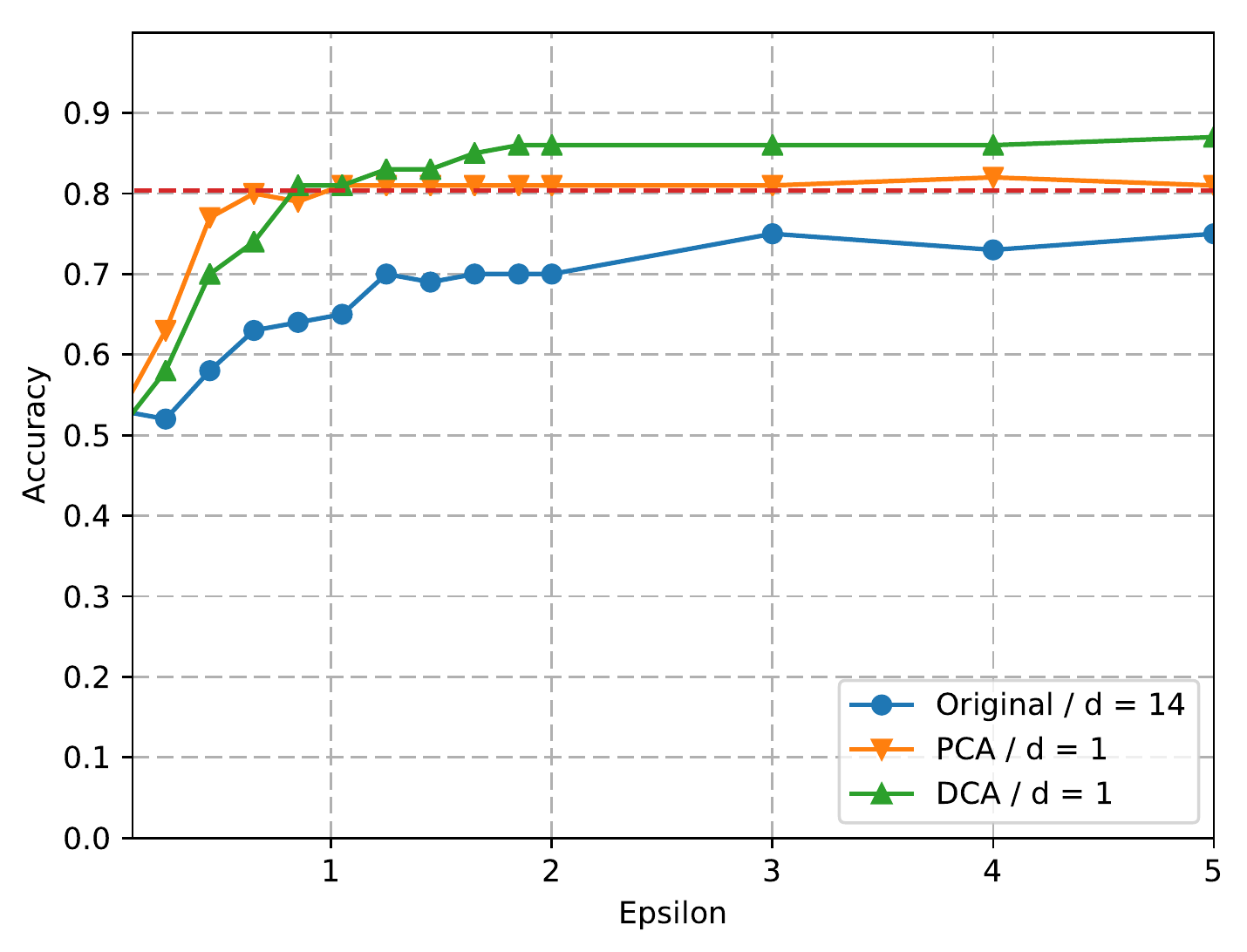}
  \caption{Australian dataset / Approach 2}
  \label{fig:australian_basicAll}
	\end{subfigure}%
	\begin{subfigure}{.33\textwidth}
  \centering
  \includegraphics[trim=0 0 0 0,clip,width=\linewidth]{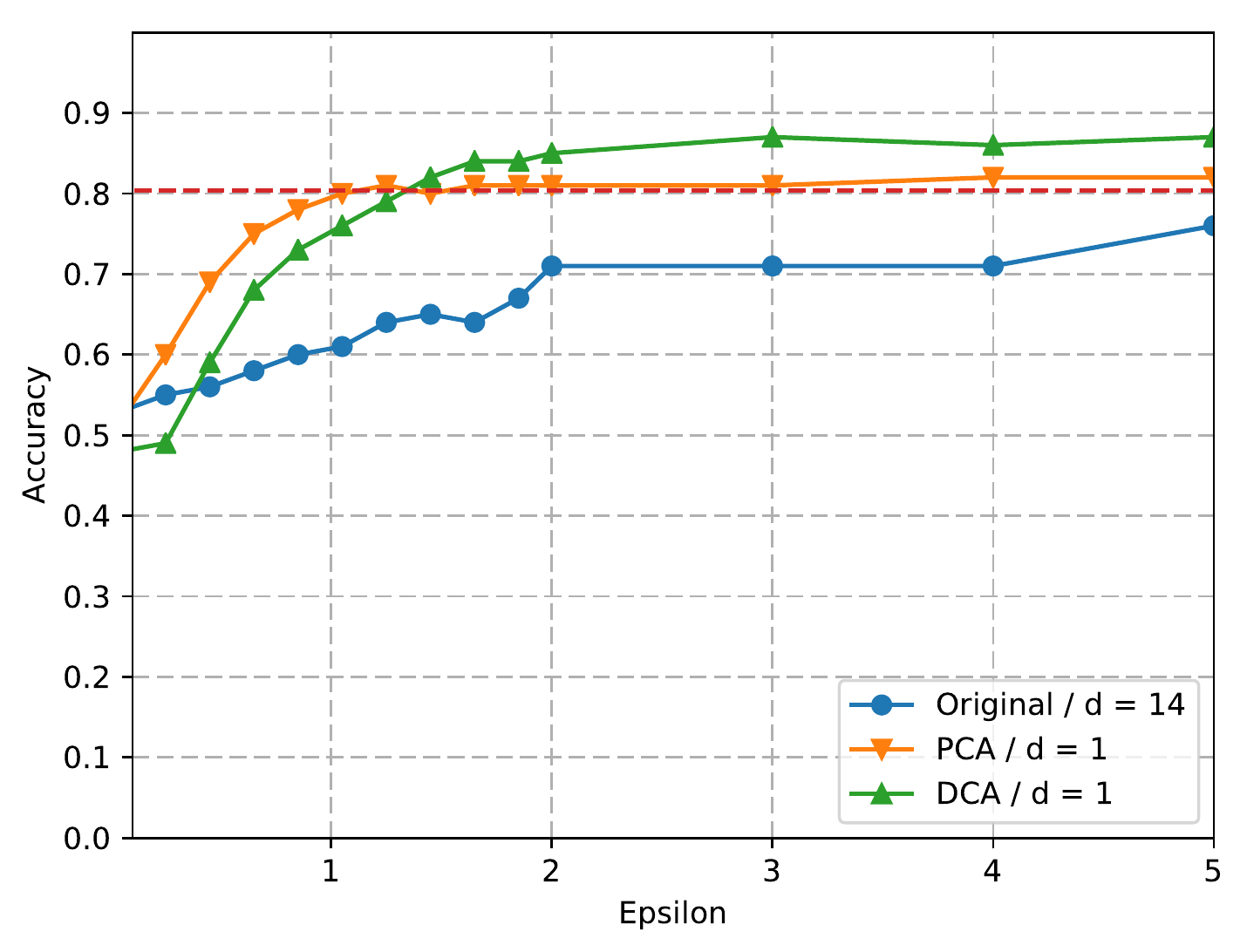}
  \caption{Australian dataset / Approach 3}
  \label{fig:australian_alg2}
\end{subfigure}%

\begin{subfigure}{.33\textwidth}
  \centering
  \includegraphics[trim=0 0 0 0,clip,width=\linewidth]{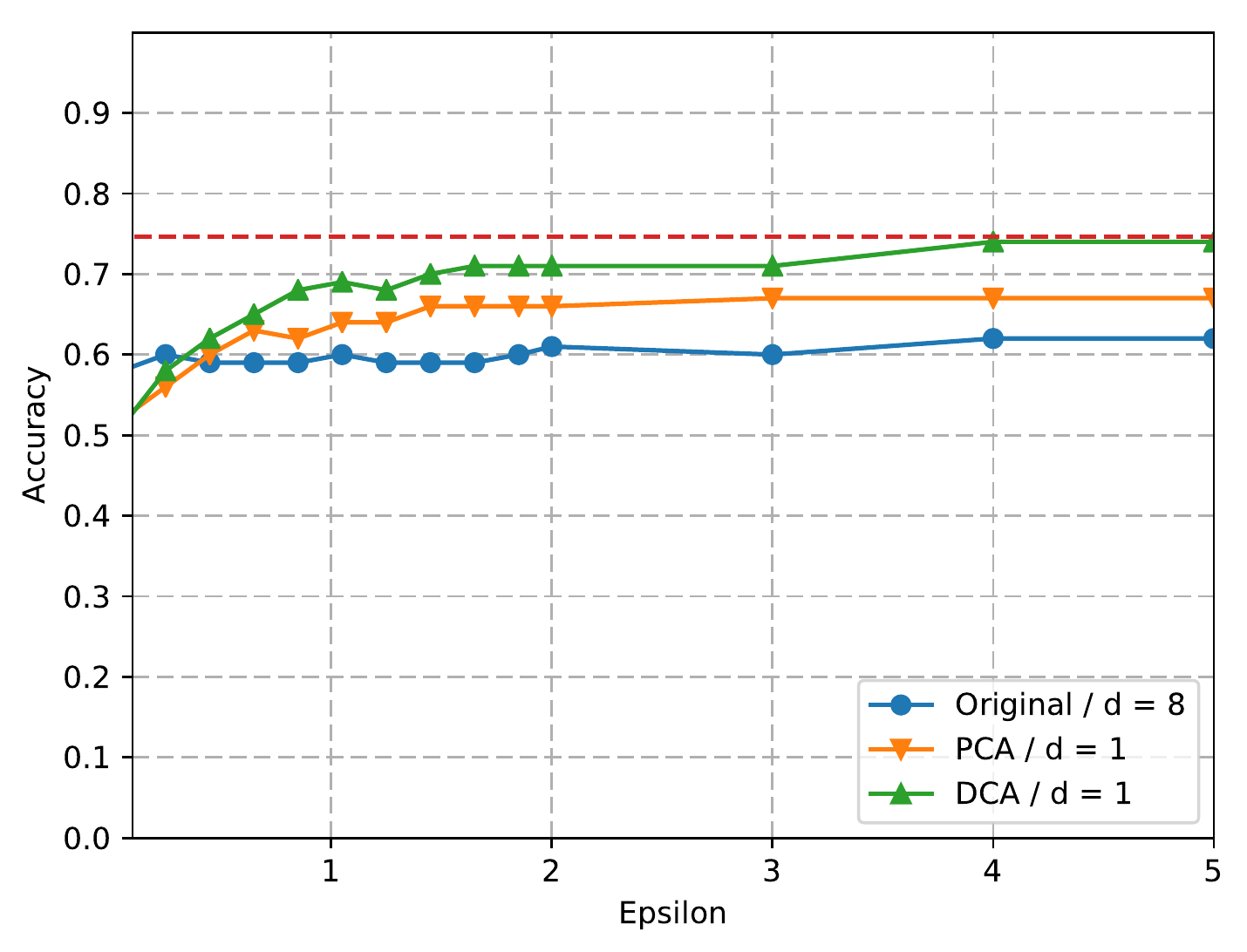}
  \caption{Diabetes dataset / Approach 1}
  \label{fig:diabetes_basicOne}
\end{subfigure}%
\begin{subfigure}{.33\textwidth}
  \centering
  \includegraphics[trim=0 0 0 0,clip,width=\linewidth]{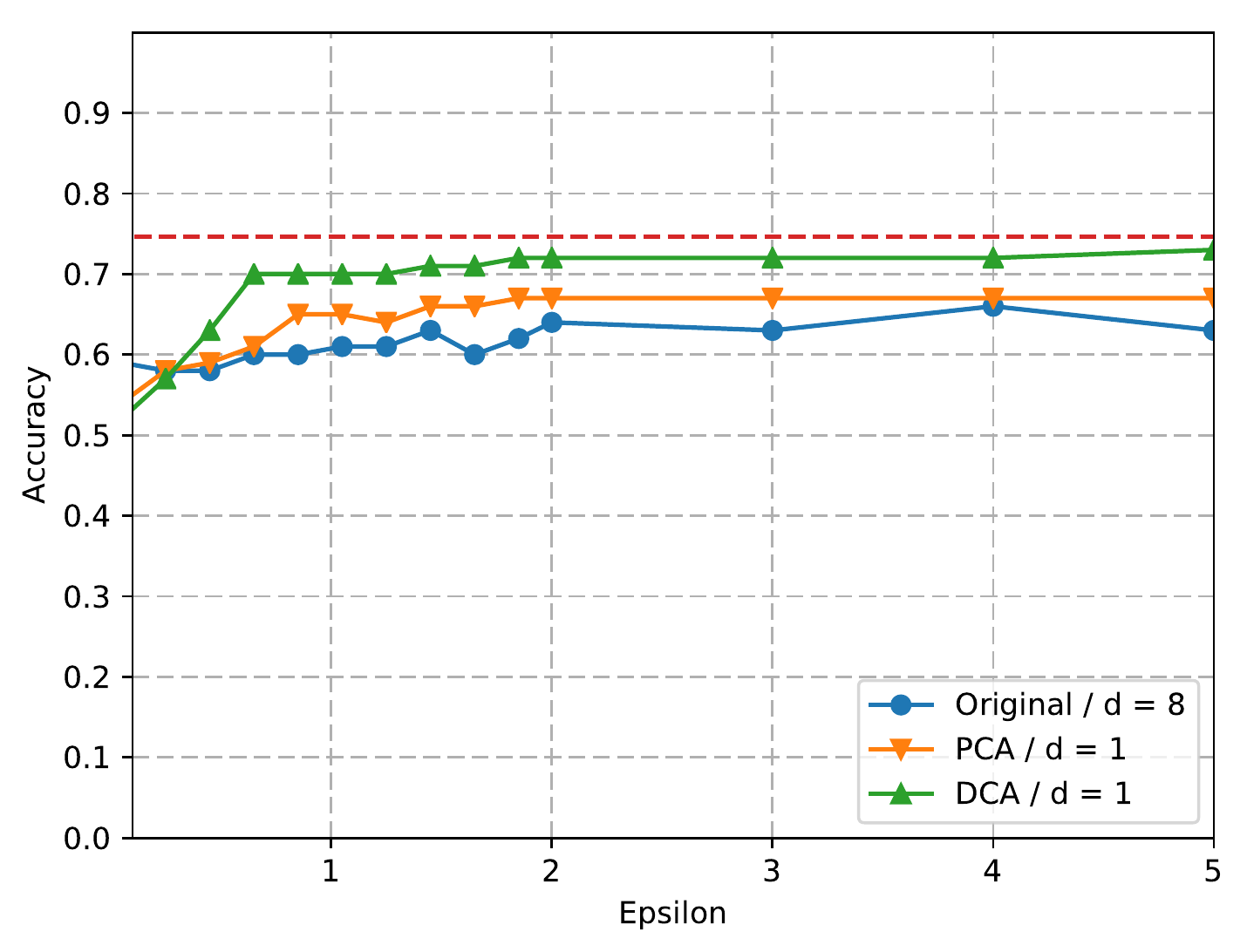}
  \caption{Diabetes dataset / Approach 2}
  \label{fig:diabetes_basicAll}
\end{subfigure}%
\begin{subfigure}{.33\textwidth}
  \centering
  \includegraphics[trim=0 0 0 0,clip,width=\linewidth]{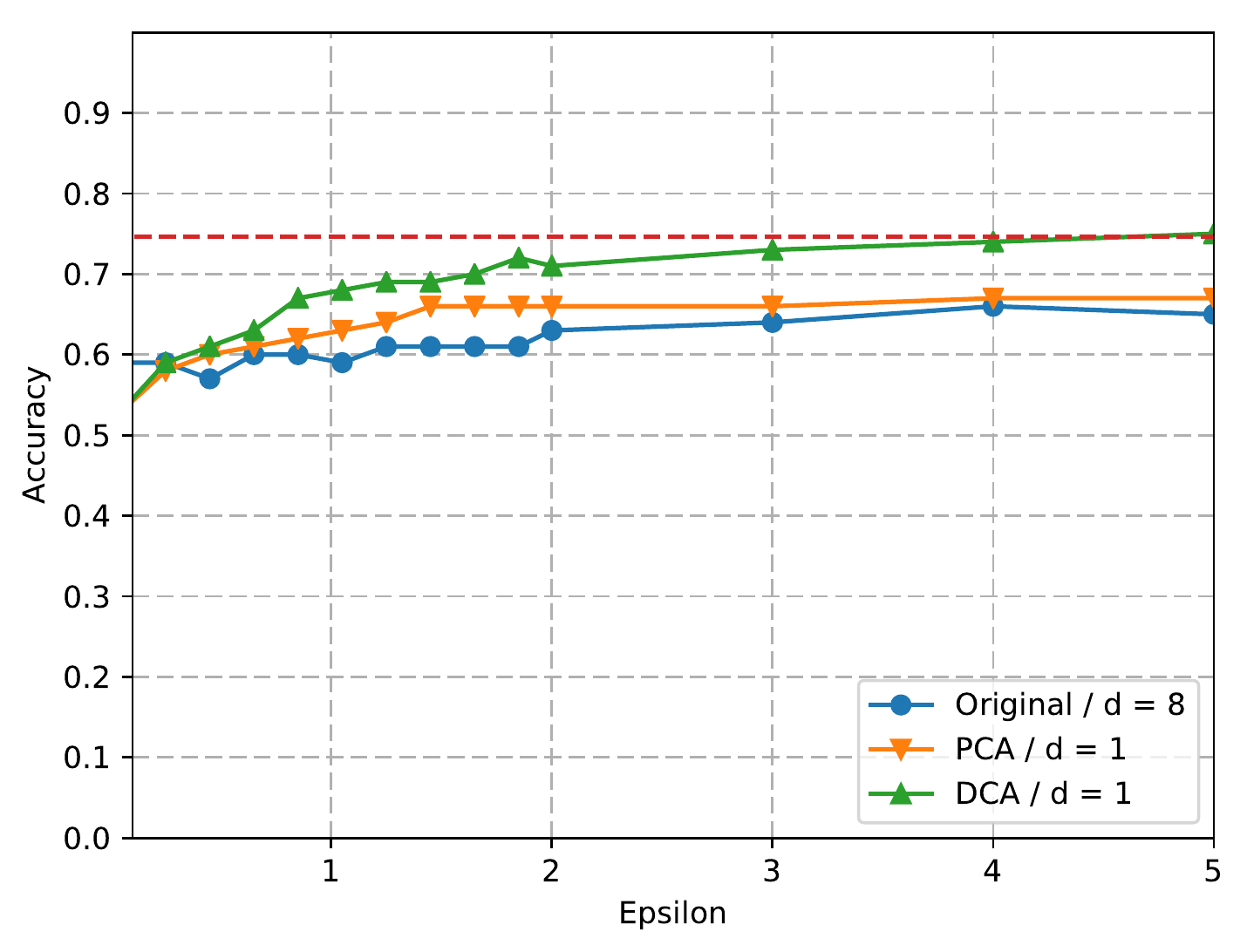}
  \caption{Diabetes dataset / Approach 3}
  \label{fig:diabetes_alg2}
\end{subfigure}%
\vspace{-3mm}
\caption{Classification accuracy for datasets with continuous features using Gaussian Naive Bayes}
\vspace{-3mm}
\label{fig:ldp-gnb}
\end{figure*}

We also applied locally differentially private Gaussian Naive Bayes (LDP-GNB) on the same two datasets. We implemented all three perturbation approaches for multi-dimensional data explained in Section \ref{subsubsec:ldpMulti}. Figure \ref{fig:ldp-gnb} shows the results of performing LDP-GNB on these two datasets. Among three approaches, the first one results in lowest utility since individuals report all features by adding more noise (i.e. propotional to the number of dimensions). In each figure, three curves are shown which correspond to using the original data (with 14 or 8 features for Australian and Diabetes datasets, respectively), or projecting the data using PCA or DCA before applying the LDP noise. The positive effect of reducing the dimensions can be clearly seen in all figures. In both datasets, and for PCA and DCA, the number of reduced dimensions were one. DCA or PCA always performs better than the original data, and for all perturbation approaches. 

Finally, when we compare discretization and Gaussian Naive Bayes for continuous data, it can be concluded that discretization provides better accuracy than Gaussian Naive Bayes. Especially for smaller $\epsilon$ values, the superiority of discretization is more apparent. Although it is not possible to compare the amount of noise for randomized response and Laplace mechanism, discretization possibly causes less noise due to smaller input domain.

\section{Related Work}
\label{sec:rel}

Privacy-preserving Naive Bayes classification has been studied before in different settings. Kantarcioglu et al. \cite{kantarcioglu2003privacy} proposed privacy-preserving Naive Bayes classifier for horizontally partitioned data. Their solution is secure in semi-honest threat model and utilizes computationally expensive cryptographic techniques such as oblivious transfer. Vaidya et al. \cite{vaidya2004privacy} addressed the same problem for vertically partitioned data. They also used secure multi-party computation primitives which are computationally expensive operations. Naive Bayes classification under differential privacy has been studied in \cite{vaidya2013differentially}. In \cite{vaidya2013differentially}, centralized setting for differential privacy is considered where the data owner has a training data and aims to release classifier by protecting privacy. They explain how to compute the sensitivity and add Laplace noise to satisfy differential privacy in Naive Bayes classifier. Li et al. \cite{li2018differentially} extended it to multiple data owners. Even though their problem setting is similar to our case, they guarantee the differential privacy at global level by calculating the global sensitivity and applying Laplace noise to the counts. Their solution does not satisfy the differential privacy in the local setting and preserves individual privacy with encryption techniques. Although privacy-preserving Naive Bayes classifier has been studied under different privacy settings such as horizontally or vertically partitioned data, and centralized differential privacy, none of them addresses the problem under LDP.

Most of the work in the literature about differential privacy consider the centralized setting. One of the earliest work on differential privacy in the local setting is Google's RAPPOR \cite{erlingsson2014rappor}. They proposed using randomized response mechanism to satisfy $\epsilon$-LDP and using bloom filters to decrease communication cost. Bassily et al. \cite{bassily2015local} also proposed a method to satisfy LDP in frequency estimation utilizing random matrix projection. Wang et al. \cite{wang2017locally} introduced a framework of pure LDP protocols to generalize the frequency estimation protocols in the literature and they proposed two new protocols for frequency estimation. We utilize these protocols in our work as mentioned in Section \ref{subsec:ldp}. Other than frequency estimation, some other problems such as heavy hitters \cite{bassily2017practical} and marginal release \cite{kulkarni2017marginal} have also been studied under LDP. The most similar work to our work is \cite{cyphers2017anonml}, which presents a system to do machine learning by satisfying LDP. To achieve better accuracy, they reduced the size of input domain to two and they also considered a binary classification model that has only two class labels. Using LDP frequency estimation the statistics about the features are estimated and using these statistics synthetic data is generated to train classification model. In our work, we do not especially address binary classification problem, and hence the number of class labels can be more than two. In addition, input domain for the features can have more than two values. By keeping the relationship between class labels and features, we allow estimation of probabilities for Naive Bayes classifier without a need for generating synthetic data.

\section{Conclusion}
\label{sec:conc}

We proposed methods for applying locally differentially private frequency and statistics estimation protocols to collect training data in Naive Bayes classification. Using the proposed methods, one can estimate all necessary probabilities to be used in Naive Bayes classification for both discrete and continuous data. To be able to estimate the conditional probabilities, the proposed methods preserve the relationship between features and class labels during the selection of inputs. Our experiment results indicate that the classification accuracy of LDP Naive Bayes for $\epsilon > 2$ is very close to the accuracy without privacy. Even for smaller $\epsilon$ values, the accuracy is remarkable when Direct Encoding or Unary Encoding schemes are used for discrete data and when discretization is used for continuous data. In addition, experiment results show that using dimensionality reduction techniques such as DCA improves the accuracy of the proposed methods for continuous data. The proposed methods facilitate collecting large training data to use in Naive Bayes classifier without compromising the privacy of the individuals providing training data. Other than Naive Bayes, LDP techniques can be utilized in different machine learning methods which can be considered as potential future work.

%
\bibliographystyle{ACM-Reference-Format}
\bibliography{sample-base}


\begin{thebibliography}{19}


\ifx \showCODEN    \undefined \def \showCODEN     #1{\unskip}     \fi
\ifx \showDOI      \undefined \def \showDOI       #1{#1}\fi
\ifx \showISBNx    \undefined \def \showISBNx     #1{\unskip}     \fi
\ifx \showISBNxiii \undefined \def \showISBNxiii  #1{\unskip}     \fi
\ifx \showISSN     \undefined \def \showISSN      #1{\unskip}     \fi
\ifx \showLCCN     \undefined \def \showLCCN      #1{\unskip}     \fi
\ifx \shownote     \undefined \def \shownote      #1{#1}          \fi
\ifx \showarticletitle \undefined \def \showarticletitle #1{#1}   \fi
\ifx \showURL      \undefined \def \showURL       {\relax}        \fi
\providecommand\bibfield[2]{#2}
\providecommand\bibinfo[2]{#2}
\providecommand\natexlab[1]{#1}
\providecommand\showeprint[2][]{arXiv:#2}

\bibitem[\protect\citeauthoryear{Bassily, Nissim, Stemmer, and
  Thakurta}{Bassily et~al\mbox{.}}{2017}]%
        {bassily2017practical}
\bibfield{author}{\bibinfo{person}{Raef Bassily}, \bibinfo{person}{Kobbi
  Nissim}, \bibinfo{person}{Uri Stemmer}, {and} \bibinfo{person}{Abhradeep~Guha
  Thakurta}.} \bibinfo{year}{2017}\natexlab{}.
\newblock \showarticletitle{Practical locally private heavy hitters}. In
  \bibinfo{booktitle}{\emph{Advances in Neural Information Processing
  Systems}}. \bibinfo{pages}{2288--2296}.
\newblock


\bibitem[\protect\citeauthoryear{Bassily and Smith}{Bassily and Smith}{2015}]%
        {bassily2015local}
\bibfield{author}{\bibinfo{person}{Raef Bassily} {and} \bibinfo{person}{Adam
  Smith}.} \bibinfo{year}{2015}\natexlab{}.
\newblock \showarticletitle{Local, private, efficient protocols for succinct
  histograms}. In \bibinfo{booktitle}{\emph{Proceedings of the forty-seventh
  annual ACM symposium on Theory of computing}}. ACM,
  \bibinfo{pages}{127--135}.
\newblock


\bibitem[\protect\citeauthoryear{Chaudhuri, Monteleoni, and Sarwate}{Chaudhuri
  et~al\mbox{.}}{2011}]%
        {chaudhuri2011differentially}
\bibfield{author}{\bibinfo{person}{Kamalika Chaudhuri}, \bibinfo{person}{Claire
  Monteleoni}, {and} \bibinfo{person}{Anand~D Sarwate}.}
  \bibinfo{year}{2011}\natexlab{}.
\newblock \showarticletitle{Differentially private empirical risk
  minimization}.
\newblock \bibinfo{journal}{\emph{Journal of Machine Learning Research}}
  \bibinfo{volume}{12}, \bibinfo{number}{Mar} (\bibinfo{year}{2011}),
  \bibinfo{pages}{1069--1109}.
\newblock


\bibitem[\protect\citeauthoryear{Cormode, Kulkarni, and Srivastava}{Cormode
  et~al\mbox{.}}{2018}]%
        {kulkarni2017marginal}
\bibfield{author}{\bibinfo{person}{Graham Cormode}, \bibinfo{person}{Tejas
  Kulkarni}, {and} \bibinfo{person}{Divesh Srivastava}.}
  \bibinfo{year}{2018}\natexlab{}.
\newblock \showarticletitle{Marginal release under local differential privacy}.
  In \bibinfo{booktitle}{\emph{Proceedings of the 2018 International Conference
  on Management of Data}}. ACM, \bibinfo{pages}{131--146}.
\newblock


\bibitem[\protect\citeauthoryear{Cyphers and Veeramachaneni}{Cyphers and
  Veeramachaneni}{2017}]%
        {cyphers2017anonml}
\bibfield{author}{\bibinfo{person}{Bennett Cyphers} {and}
  \bibinfo{person}{Kalyan Veeramachaneni}.} \bibinfo{year}{2017}\natexlab{}.
\newblock \showarticletitle{AnonML: Locally private machine learning over a
  network of peers}. In \bibinfo{booktitle}{\emph{Data Science and Advanced
  Analytics (DSAA), 2017 IEEE International Conference on}}. IEEE,
  \bibinfo{pages}{549--560}.
\newblock


\bibitem[\protect\citeauthoryear{Dheeru and Karra~Taniskidou}{Dheeru and
  Karra~Taniskidou}{2017}]%
        {Dua:2017}
\bibfield{author}{\bibinfo{person}{Dua Dheeru} {and} \bibinfo{person}{Efi
  Karra~Taniskidou}.} \bibinfo{year}{2017}\natexlab{}.
\newblock \bibinfo{title}{{UCI} Machine Learning Repository}.
\newblock
\newblock
\urldef\tempurl%
\url{http://archive.ics.uci.edu/ml}
\showURL{%
\tempurl}


\bibitem[\protect\citeauthoryear{Dwork}{Dwork}{2008}]%
        {dwork2008differential}
\bibfield{author}{\bibinfo{person}{Cynthia Dwork}.}
  \bibinfo{year}{2008}\natexlab{}.
\newblock \showarticletitle{Differential privacy: A survey of results}. In
  \bibinfo{booktitle}{\emph{International Conference on Theory and Applications
  of Models of Computation}}. Springer, \bibinfo{pages}{1--19}.
\newblock


\bibitem[\protect\citeauthoryear{Erlingsson, Pihur, and Korolova}{Erlingsson
  et~al\mbox{.}}{2014}]%
        {erlingsson2014rappor}
\bibfield{author}{\bibinfo{person}{{\'U}lfar Erlingsson},
  \bibinfo{person}{Vasyl Pihur}, {and} \bibinfo{person}{Aleksandra Korolova}.}
  \bibinfo{year}{2014}\natexlab{}.
\newblock \showarticletitle{Rappor: Randomized aggregatable privacy-preserving
  ordinal response}. In \bibinfo{booktitle}{\emph{Proceedings of the 2014 ACM
  SIGSAC conference on computer and communications security}}. ACM,
  \bibinfo{pages}{1054--1067}.
\newblock


\bibitem[\protect\citeauthoryear{Jagannathan, Pillaipakkamnatt, and
  Wright}{Jagannathan et~al\mbox{.}}{2009}]%
        {jagannathan2009practical}
\bibfield{author}{\bibinfo{person}{Geetha Jagannathan},
  \bibinfo{person}{Krishnan Pillaipakkamnatt}, {and} \bibinfo{person}{Rebecca~N
  Wright}.} \bibinfo{year}{2009}\natexlab{}.
\newblock \showarticletitle{A practical differentially private random decision
  tree classifier}. In \bibinfo{booktitle}{\emph{Data Mining Workshops, 2009.
  ICDMW'09. IEEE International Conference on}}. IEEE,
  \bibinfo{pages}{114--121}.
\newblock


\bibitem[\protect\citeauthoryear{Kairouz, Oh, and Viswanath}{Kairouz
  et~al\mbox{.}}{2014}]%
        {kairouz2014extremal}
\bibfield{author}{\bibinfo{person}{Peter Kairouz}, \bibinfo{person}{Sewoong
  Oh}, {and} \bibinfo{person}{Pramod Viswanath}.}
  \bibinfo{year}{2014}\natexlab{}.
\newblock \showarticletitle{Extremal mechanisms for local differential
  privacy}. In \bibinfo{booktitle}{\emph{Advances in neural information
  processing systems}}. \bibinfo{pages}{2879--2887}.
\newblock


\bibitem[\protect\citeauthoryear{Kantarc{\i}oglu, Vaidya, and
  Clifton}{Kantarc{\i}oglu et~al\mbox{.}}{2003}]%
        {kantarcioglu2003privacy}
\bibfield{author}{\bibinfo{person}{Murat Kantarc{\i}oglu},
  \bibinfo{person}{Jaideep Vaidya}, {and} \bibinfo{person}{C Clifton}.}
  \bibinfo{year}{2003}\natexlab{}.
\newblock \showarticletitle{Privacy preserving naive bayes classifier for
  horizontally partitioned data}. In \bibinfo{booktitle}{\emph{IEEE ICDM
  workshop on privacy preserving data mining}}. \bibinfo{pages}{3--9}.
\newblock


\bibitem[\protect\citeauthoryear{Kung}{Kung}{2014}]%
        {kung2014kernel}
\bibfield{author}{\bibinfo{person}{Sun~Yuan Kung}.}
  \bibinfo{year}{2014}\natexlab{}.
\newblock \bibinfo{booktitle}{\emph{Kernel methods and machine learning}}.
\newblock \bibinfo{publisher}{Cambridge University Press}.
\newblock


\bibitem[\protect\citeauthoryear{Li, Li, Liu, Li, and Jia}{Li
  et~al\mbox{.}}{2018}]%
        {li2018differentially}
\bibfield{author}{\bibinfo{person}{Tong Li}, \bibinfo{person}{Jin Li},
  \bibinfo{person}{Zheli Liu}, \bibinfo{person}{Ping Li}, {and}
  \bibinfo{person}{Chunfu Jia}.} \bibinfo{year}{2018}\natexlab{}.
\newblock \showarticletitle{Differentially private naive bayes learning over
  multiple data sources}.
\newblock \bibinfo{journal}{\emph{Information Sciences}}  \bibinfo{volume}{444}
  (\bibinfo{year}{2018}), \bibinfo{pages}{89--104}.
\newblock


\bibitem[\protect\citeauthoryear{Nguy{\^e}n, Xiao, Yang, Hui, Shin, and
  Shin}{Nguy{\^e}n et~al\mbox{.}}{2016}]%
        {nguyen2016collecting}
\bibfield{author}{\bibinfo{person}{Th{\^o}ng~T Nguy{\^e}n},
  \bibinfo{person}{Xiaokui Xiao}, \bibinfo{person}{Yin Yang},
  \bibinfo{person}{Siu~Cheung Hui}, \bibinfo{person}{Hyejin Shin}, {and}
  \bibinfo{person}{Junbum Shin}.} \bibinfo{year}{2016}\natexlab{}.
\newblock \showarticletitle{Collecting and analyzing data from smart device
  users with local differential privacy}.
\newblock \bibinfo{journal}{\emph{arXiv preprint arXiv:1606.05053}}
  (\bibinfo{year}{2016}).
\newblock


\bibitem[\protect\citeauthoryear{Qin, Yang, Yu, Khalil, Xiao, and Ren}{Qin
  et~al\mbox{.}}{2016}]%
        {qin2016heavy}
\bibfield{author}{\bibinfo{person}{Zhan Qin}, \bibinfo{person}{Yin Yang},
  \bibinfo{person}{Ting Yu}, \bibinfo{person}{Issa Khalil},
  \bibinfo{person}{Xiaokui Xiao}, {and} \bibinfo{person}{Kui Ren}.}
  \bibinfo{year}{2016}\natexlab{}.
\newblock \showarticletitle{Heavy hitter estimation over set-valued data with
  local differential privacy}. In \bibinfo{booktitle}{\emph{Proceedings of the
  2016 ACM SIGSAC Conference on Computer and Communications Security}}. ACM,
  \bibinfo{pages}{192--203}.
\newblock


\bibitem[\protect\citeauthoryear{Rubinstein, Bartlett, Huang, and
  Taft}{Rubinstein et~al\mbox{.}}{2012}]%
        {rubinstein2012learning}
\bibfield{author}{\bibinfo{person}{Benjamin~IP Rubinstein},
  \bibinfo{person}{Peter~L Bartlett}, \bibinfo{person}{Ling Huang}, {and}
  \bibinfo{person}{Nina Taft}.} \bibinfo{year}{2012}\natexlab{}.
\newblock \showarticletitle{Learning in a Large Function Space:
  Privacy-Preserving Mechanisms for SVM Learning}.
\newblock \bibinfo{journal}{\emph{Journal of Privacy and Confidentiality}}
  \bibinfo{volume}{4}, \bibinfo{number}{1} (\bibinfo{year}{2012}),
  \bibinfo{pages}{65--100}.
\newblock


\bibitem[\protect\citeauthoryear{Vaidya and Clifton}{Vaidya and
  Clifton}{2004}]%
        {vaidya2004privacy}
\bibfield{author}{\bibinfo{person}{Jaideep Vaidya} {and} \bibinfo{person}{Chris
  Clifton}.} \bibinfo{year}{2004}\natexlab{}.
\newblock \showarticletitle{Privacy preserving naive bayes classifier for
  vertically partitioned data}. In \bibinfo{booktitle}{\emph{Proceedings of the
  2004 SIAM International Conference on Data Mining}}. SIAM,
  \bibinfo{pages}{522--526}.
\newblock


\bibitem[\protect\citeauthoryear{Vaidya, Shafiq, Basu, and Hong}{Vaidya
  et~al\mbox{.}}{2013}]%
        {vaidya2013differentially}
\bibfield{author}{\bibinfo{person}{Jaideep Vaidya}, \bibinfo{person}{Basit
  Shafiq}, \bibinfo{person}{Anirban Basu}, {and} \bibinfo{person}{Yuan Hong}.}
  \bibinfo{year}{2013}\natexlab{}.
\newblock \showarticletitle{Differentially private naive bayes classification}.
  In \bibinfo{booktitle}{\emph{Web Intelligence (WI) and Intelligent Agent
  Technologies (IAT), 2013 IEEE/WIC/ACM International Joint Conferences on}},
  Vol.~\bibinfo{volume}{1}. IEEE, \bibinfo{pages}{571--576}.
\newblock


\bibitem[\protect\citeauthoryear{Wang, Blocki, Li, and Jha}{Wang
  et~al\mbox{.}}{2017}]%
        {wang2017locally}
\bibfield{author}{\bibinfo{person}{Tianhao Wang}, \bibinfo{person}{Jeremiah
  Blocki}, \bibinfo{person}{Ninghui Li}, {and} \bibinfo{person}{Somesh Jha}.}
  \bibinfo{year}{2017}\natexlab{}.
\newblock \showarticletitle{Locally differentially private protocols for
  frequency estimation}. In \bibinfo{booktitle}{\emph{Proc. of the 26th USENIX
  Security Symposium}}. \bibinfo{pages}{729--745}.
\newblock


\end{thebibliography}

%
%
%
%
%
%
%
%
%

\end{document}